\documentclass[letterpaper]{article} 
\usepackage{main}  
\usepackage{times}  
\usepackage{helvet}  
\usepackage{courier}  
\usepackage[hyphens]{url}  
\usepackage{graphicx} 
\usepackage{natbib}  
\usepackage{caption} 
\usepackage{algorithm}
\usepackage{algorithmic}
\usepackage{multirow}
\usepackage{amsmath}
\usepackage{booktabs}
\usepackage{amssymb}
\usepackage{subfigure}
\usepackage{colortbl}
\usepackage[dvipsnames]{xcolor}
\newcommand{\jf}[1]{\textcolor{black}{#1}}
\usepackage{newfloat}
\usepackage{listings}
\DeclareCaptionStyle{ruled}{labelfont=normalfont,labelsep=colon,strut=off} 
\floatstyle{ruled}
\newfloat{listing}{tb}{lst}{}
\floatname{listing}{Listing}
\title{Skating-Mixer: Long-Term Sport Audio-Visual Modeling with MLPs}
\author{
    Jingfei Xia\thanks{The first two authors contributed equally: work is done when they worked as visiting scholars in SUSTech.}\textsuperscript{\rm 1,2},
    Mingchen Zhuge\footnotemark[1]\textsuperscript{\rm 1,3}, \\
    Tiantian Geng\textsuperscript{\rm 1}, 
    Shun Fan\textsuperscript{\rm 1}, 
    Yuantai Wei\textsuperscript{\rm 1}, 
    Zhenyu He\textsuperscript{\rm 4}, 
    Feng Zheng\thanks{Corresponding author}\textsuperscript{\rm 1}
}
\affiliations {
    \textsuperscript{\rm 1}Southern University of Science and Technology 
    \textsuperscript{\rm 2}The Chinese University of Hong Kong \\
    \textsuperscript{\rm 3}AI Initiative, KAUST 
    \textsuperscript{\rm 4}Harbin Institute of Technology (Shenzhen)\\
    xj022@ie.cuhk.edu.hk, mingchen.zhuge@kaust.edu.sa, zhenyuhe@hit.edu.cn,\\
    \{gengtiantian97, 27957322s, santyelegy, zfeng02\}@gmail.com
}

\usepackage{bibentry}

\begin{document}

\maketitle


\begin{abstract}
Figure skating scoring is challenging because it requires judging the technical moves of the players as well as their coordination with the background music.
Most learning-based methods cannot solve it well for two reasons: 1) each move in figure skating changes quickly, hence simply applying traditional frame sampling will lose a lot of valuable information, especially in $3$ to $5$ minutes long videos; 
2) prior methods rarely considered the critical audio-visual relationship in their models.
Due to these reasons, we introduce a 
novel architecture, named \textbf{Skating-Mixer}.
It extends the 
MLP framework into a multimodal fashion and effectively learns long-term representations through our designed memory recurrent unit (MRU).
Aside from the model, we 
collected a high-quality audio-visual \textbf{FS1000} dataset, which contains over 1000 videos on 8 types of programs with 7 different rating metrics, overtaking other datasets in both quantity and diversity.
%
Experiments show the proposed method achieves SOTAs over all major metrics on the public Fis-V and our FS1000 dataset. 
In addition, we include an analysis applying our method to the recent competitions in Beijing 2022 Winter Olympic Games, proving our method has strong applicability. 
\end{abstract}

\section{Introduction}
  
\begin{figure*}[h]
\begin{center}
\includegraphics[width=0.75 \linewidth]{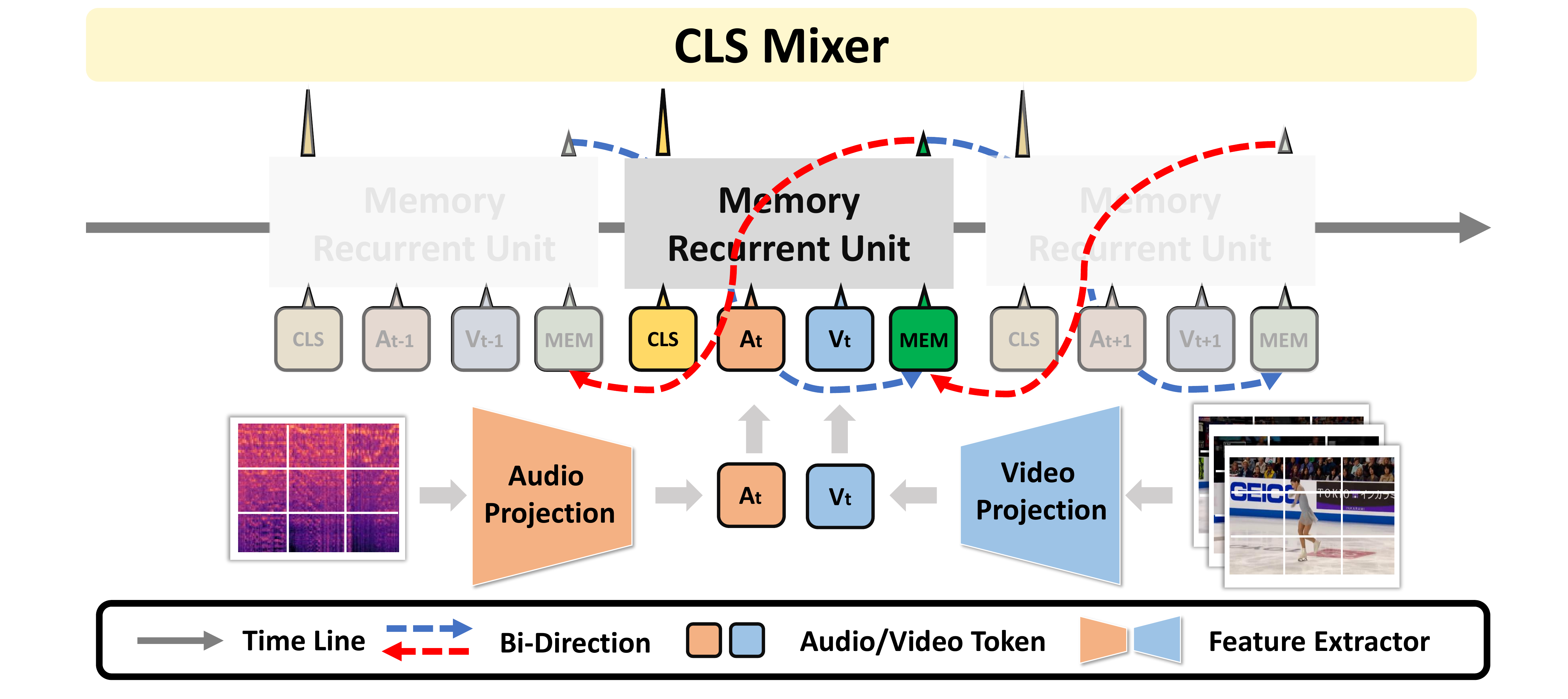}
\end{center}
\vspace{-10pt}
    \caption{\textbf{\textbf{Pipeline of Skating-Mixer.}} We utilize patch modeling methods just like \cite{dosovitskiy2020image,tolstikhin2021mlp}, and use TimeSformer~\cite{bertasius2021spacetime} and AST~\cite{gong2021ast} as our projection backbones. 
    The memory recurrent unit (MRU) of Skating-Mixer works for learning sequential temporal multimodal information.    
    After integral learning in both spatial and temporal information, Skating-Mixer obtains a representation of long-range video. 
    We integrate the output of the CLS Mixer and the output of \texttt{[MEM]} from the last clip to finish the scoring.
    \texttt{[MEM]} denotes the memory token and \texttt{[CLS]} denotes the class token.}
\label{fig:pipeline}
\vspace{-10pt}
\end{figure*}

Due to the importance of fair competition, many worldwide sports committees devote themselves to regularizing the behaviors of both athletes and referees. 
As a supplementary tool, crowd workers seek to employ objective machine intelligence to judge the performances in competitions. Therefore, many learning-based assessment models~\cite{pirsiavash2014assessing,parmar2017learning,pan2019action,jain2020action,xiang2018s3d,li2018end} have been introduced in recent years.
\jf{However, few works proposed to score figure skating videos due to several key challenges:}

\begin{itemize}
    \item[$\bullet$]  
    \textbf{Requiring strong video representation learning.}
    (1) Figure skating videos are 3$-$5 minutes long and contain manifold technical movements, requiring effective representation learning on the long videos with large fps (frame per second) \jf{which results in large size of inputs.} (2) Both audio and video should be considered when calculating the scores in figure skating.
    \item[$\bullet$] 
    
    \textbf{Missing high-quality dataset. }
    Unlike common videos,
    figure skating videos are sourced from live sporting tournaments that require extensive manual process effort. This could be the reason that existing datasets~\cite{xu2019learning,liu2020fsd} are not comprehensive enough (in scale or diversity) to cover figure skating.  
\end{itemize}

To solve the first challenge, earlier work~\cite{liu2020fsd} utilized a hierarchical LSTM model to capture the local and global information in figure skating videos. The recent Eagle-eye method~\cite{nekoui2021eagle} considers pose heatmaps and appearance features jointly. Although these two methods achieve comparably good results, they remain obvious deficiencies. 
Firstly, their methods can hardly generalize to different categories of competitions in the real scene.
More importantly, they merely consider visual modality. 
While technical action (visual modality) is important in figure skating, we argue that background music (audio modality) should not be ignored as well.
An efficient model which considers both visual and audio cues is urgently needed in this field.

Convolutional Neural Networks~\cite{cnn} have long been used for various computer vision tasks~\cite{zhuge2022cubenet}. With the success in Natural Language Processing, Transformers~\cite{vaswani2017attention} have been introduced into the computer vision area and show great power in multimodal learning compared with CNNs~\cite{zhuge2021kaleido}. However, the complexity of Vision Transformer~\cite{dosovitskiy2020image} is quadratic in the
number of input patches and it usually requires a large amount of data to train~\cite{Liu2021EfficientTO}, which is hardly satisfied in our specific task. Recently, a pure MLP-based structure MLP-Mixer has been proposed and shows promising results on the image classification task~\cite{tolstikhin2021mlp} with linear complexity. Yet, this simple structure has not been deeply explored in the audio-visual area.
Therefore, we introduce Skating-Mixer, which is the pioneer of MLP-based multimodal architecture and also the first to 
score figure skating with both auditory and visual feature.
Skating-Mixer has the following properties: $(1)$ It simultaneously models audio and visual features, and learns the long-term joint representation in an effective way; $(2)$ By using the memory recurrent unit (MRU), this approach accurately predicts the results using extremely long-range cues; $(3)$ 
With its simple design, it could avoid the gradient vanishing and exploding problems in the vanilla recurrent neural network~\cite{rnn}.

Moreover, we observe that few high-quality datasets are set up for this task.
Fis-V~\cite{xu2019learning} only contains ladies short program videos.
Since the data distribution is monotonous, this dataset is less challenging.
In this case, we build a new dataset FS1000 with more than 1000 videos, 8 categories of figure skating, and 7 detailed scores to increase the diversity and quantity. The dataset requires the model to learn the underlying features across different figure skating videos.
Experiments conducted on both datasets demonstrate that Skating-Mixer not only achieves state-of-the-art results 
but also has the ability to generalize to all sorts of figure skating, making a good example  to tackle multimodal representation in sports.
In summary, the primary contributions of this paper are listed as following:
\begin{itemize}
    \item[$\bullet$]  
    \vspace{-3pt}
    We present the pioneer MLP-based multimodal framework that can model extremely long-range videos. It automatically scores figure skating with both auditory and visual information.
    \vspace{-3pt}
    \item[$\bullet$]  
    We collect a comprehensive FS1000 dataset, including more long-range videos that contain all types of figure skating with more detailed score records.
    \vspace{-2pt}    
    \item[$\bullet$]  
    We benchmark recent methods in this field on both Fis-V and FS1000 datasets.
    The proposed MLP-based and framework outperforms other CNN-based~\cite{parmar2017learning, parmar2019action}, LSTM-based~\cite{xu2019learning}, and Transformer-based~\cite{lee2020parameter} methods.
\end{itemize}

\section{Dataset}
\begin{figure*}[h]
\begin{center}
\includegraphics[width=0.9\linewidth]{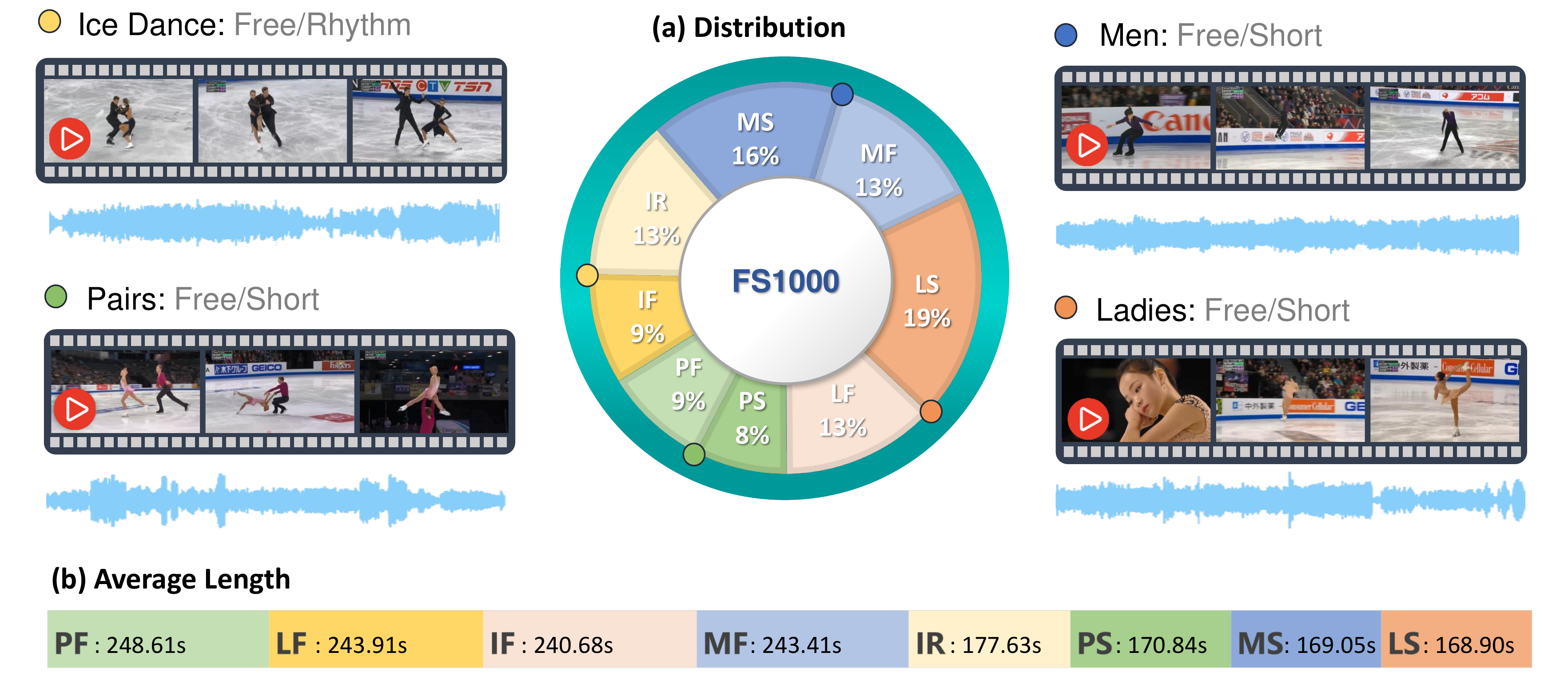}
\end{center}
\vspace{-15pt}
    \caption{
    The distribution, average length, and video samples for each category in FS1000. \textbf{{MS}:} men short program \emph{(16\%)}, \textbf{{MF}:} men free skating \emph{(13\%)}, \textbf{{LS}:} ladies short program \emph{(19\%)}, \textbf{{LF}:} ladies free skating \emph{(13\%)}, \textbf{{PS}:} pairs short program \emph{(8\%)}, \textbf{{PF}:} pairs free skating \emph{(9\%)}, \textbf{{IF}:} ice dance free skating \emph{(9\%)}, and \textbf{{IR}:} ice dance rhythm dance \emph{(13\%)}.}
\label{fig:dataset_example}
\vspace{-15pt}
\end{figure*}
To further facilitate the research of learning-based figure skating scoring, we present the largest figure skating dataset FS1000 with high-quality videos.
All videos come from high-standard international figure skating competitions and were captured by professional camera devices.
The dataset is designed for predicting scores in figure skating competitions, with rich annotations like player ID and program categories, which may facilitate this field even more. 

\subsection{Data Collection}
\noindent\textbf{Data Source.}
%
With an aim to obtain high-quality figure skating videos, we carefully selected and downloaded videos only from the top-tier international skating competitions.
Besides, we also collected and checked scores from referee reports to annotate our dataset in authoritative.\footnote{https://www.isu.org/figure-skating/entries-results/fsk-results}
Specifically, ISU World Figure Skating Championships, ISU Grand Prix of Figure Skating, \emph{etc.,} 
are selected as our main data sources.
Normally, figure skating consists of 4 primary categories: mens singles, ladies singles, pair skating and ice dance.
Each primary category contains short program and free skating (in ice dance, they are called rhythm dance and free dance). So, there are 8 subdivided categories, as shown in Figure \ref{fig:dataset_example}.


\noindent{{\bf Pre-processing.}}
The raw videos collected from figure skating competitions are usually untrimmed and record the whole procedure ranging from $1$ to $5$ hours. It consists of the performances of all players as well as highlight replay, warming up parts, and waiting for score parts. 
These redundant contents are usually not helpful for score judgment.
We initially downloaded over {$400$-hour} videos and then
manually processed all videos, only reserving pure competition performance clips of players from the exact beginning to the ending moment of background music. 

\vspace{-5pt}
\subsection{Annotation and statistics}
{\bf Annotation.} As mentioned above, there are totally eight categories of figure skating competitions, namely, men/ladies/pairs short program (\textbf{MS}, \textbf{LS}, \textbf{PS}), men/ladies/pairs free skating (\textbf{MF}, \textbf{LF}, \textbf{PF}) and ice dance rhythm dance/free dance (\textbf{IR}, \textbf{IF}). 
We carefully labeled each video with its official scores according to the referee reports, and also player ID and corresponding category. 
In the scoring regulation of figure skating, the result can be divided into two parts: Technical Element Score (\textbf{\emph{TES}}) and Program Component Score (\textbf{\emph{PCS}}).
\textbf{\emph{TES}} evaluates the difficulty and execution of all technical movements, and \textbf{\emph{PCS}} describes the overall performance, considering five aspects: the Skating Skills  (\textbf{\emph{SS}}), Transitions  (\textbf{\emph{TR}}), Performance  (\textbf{\emph{PE}}), Composition  (\textbf{\emph{CO}}), and Interpretation of music (\textbf{\emph{IN}}).
Specifically, the Skating Skills assesses the skater's command of the blade over the ice; Transitions evaluates skaters' ability to transit between technical elements naturally; Performance shows the appeal and personality of the program; Composition reflects the choreography and the purpose to the way the program is constructed, and Interpretation is more concerned with the consistency between each movement and a corresponding beat in music.
Besides, there is a factor that indicates different weights of \textbf{\emph{PCS}} scores in different competitions.
These scores are given by nine different professional referees and calculated by the referee rules of figure skating competition.
Each video in the FS1000 dataset is annotated with these information mentioned above. 

\vspace{-2pt}
\noindent{\bf Statistics.} 
There are totally 1604 figure skating videos in our FS1000 dataset: 1247 videos are for training and validation while 357 videos from contests in 2022 are for testing, including Beijing 2022 Olympics.
Each video has $\sim$5000 frames with a frame rate of 25 and is annotated with detailed ground-truth scores.
Some example frames and the percentage of the number of each category in the FS1000 dataset are shown in Figure~\ref{fig:dataset_example}(a). 
As the videos contain complete snippets of each performance, they are relatively long with a duration ranging from 2.5 minutes to 4.3 minutes, and the average duration of all videos is about 3 minutes and 20 seconds.
The details of each category's average length are given in Figure~\ref{fig:dataset_example}(b).
We can see that compared with the duration of short program and rhythm dance (2 to 3 minutes), free skating and free dance generally have a longer duration (about 4 minutes), which is consistent with the standards of different competitions.

\vspace{-4pt}
\subsection{Comparison with other datasets}
\vspace{-7pt}
\begin{table}[h!]
	\renewcommand\tabcolsep{4.0pt}
	\centering
	\scalebox{0.75}{
        \begin{tabular}{l|c|c|c|c|c|c}
        \toprule
\rowcolor{blue!10}
        \textbf{Dataset} &\textbf{Task} & \bf $\#$ Video & \bf Length  & \bf $\#$ Score & \bf $\#$ Type & \bf Feature \\  \hline
        FSD-10 & AR & 1484 & ~10h & 1 & - & V    \\
        MCFS & AS & 271 & ~5h & - & - & V    \\ \hline
        MIT-Skate & LS & 171 & 8h  & 1  & 1 & V  \\ 
        FisV & LS & 500 & 24h & 2 & 1 & V  \\
        
        FS1000 (ours) & LS & 1604 & 91h &  7 & 8 & A+V  \\  
        \bottomrule   
        \end{tabular}
    }
\caption{Dataset comparison in figure skating area. Length refers to the total length of all videos. (AR: Action Recognition, AS: Action Segmentation, LS: Long-video Scoring.)}
\vspace{-10pt}
\label{tab: dataset}
\end{table}
\noindent Here, we show the comparison between our proposed FS1000 dataset and other existing figure skating video datasets: MCFS~\cite{DBLP:conf/aaai/LiuZLZXDZ21}, FSD-10~\cite{liu2020fsd}, FisV~\cite{xu2019learning} and MIT-Skate dataset~\cite{pirsiavash2014assessing, parmar2017learning}. 
FSD-10 and MCFS focus on the technical actions in figure skating and they only contain short clips of each action instead of the complete videos, which is not suitable for our multimodal scoring task. 
MIT-Skate, Fis-V and our proposed FS1000 are all set up for long-video scoring tasks. MIT-Skate consists of videos that happened before 2012, which is outdated because of the regulation change. 
Also, it only contains 171 ladies short program videos, which is too limited in scale. 
Therefore, we focus on the last two datasets in our work. 
We can see that our dataset is the first one to utilize audio feature in this area and overtakes the other datasets in quantity and diversity.
        

\section{Method}
\label{skating Mixer}


In this section, we comprehensively introduce our MLP-based multimodal model, the \textbf{Skating-Mixer}.
In our task, the model should be capable of handling extremely long videos with large fps, which means gigantic input audio-visual feature.
\jf{Models like ViT~\cite{dosovitskiy2020image} can hardly handle these data due to the memory limit.}
Unlike previous methods~\cite{bain2021frozen,lei2021less} that sample several frames to represent a whole video, we segment the video into multiple 5-second clips and input them to projection models~\cite{gong2021ast,bertasius2021spacetime} to obtain audio and video features respectively. 
This is because our model needs to deal with both audio and visual features and sampling frames will cause these two features unaligned.
\jf{Also, sampling frames will lose information of fast motions.}
The whole framework is shown in Figure~\ref{fig:pipeline}.
Suppose there are $T$ clips in the video.
For the $t$-th clip in the video, $\textbf{A}_t$ and $\textbf{V}_t$ denote the input audio feature and video feature, respectively.

\vspace{-5pt}
\subsubsection{Memory Recurrent Unit.}
\begin{figure}[h]
\begin{center}
\includegraphics[width=0.9\linewidth]{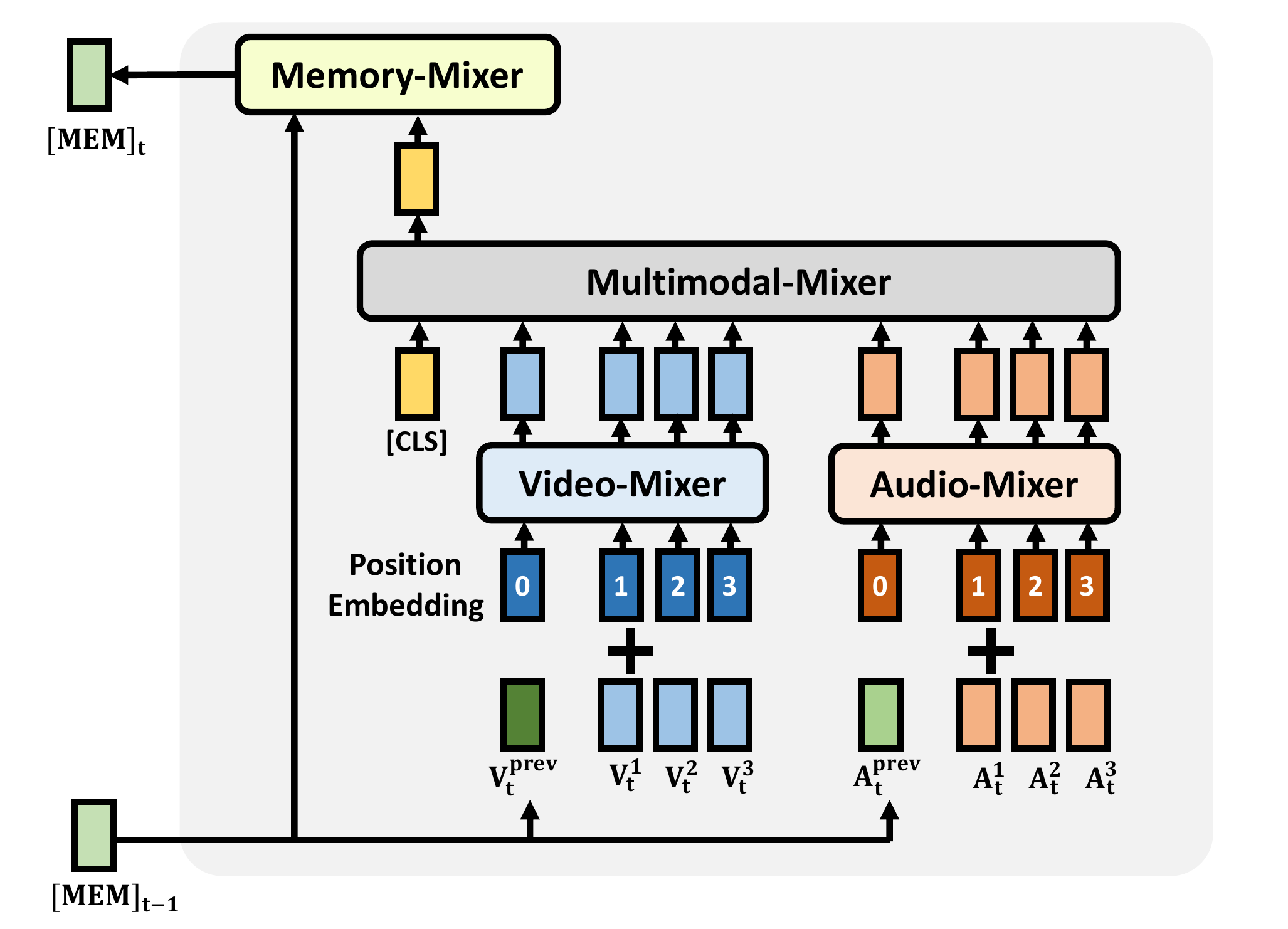}
\end{center}
\vspace{-10pt}
    \caption{
    The structure of memory recurrent unit.}
\label{fig:newstruct}
\vspace{-10pt}
\end{figure}
The structure of memory recurrent unit (MRU) is shown in Figure~\ref{fig:newstruct}. This unit tends to capture the long-term relationships in the video and obtain a comprehensive overview of the whole video. 
The main idea of MRU is to split the video into several clips and process clip by clip. However, it is also necessary to let each clip acquire the information from the previous clips.
Therefore, a memory token \texttt{[MEM]} is applied with audio and video input. The memory token is a learnable randomly-initialized vector. At the $t$-th clip, the input memory $\texttt{[MEM]}_{t-1}$ is first passed into two bottleneck structures to extract audio and video-related features from the previous clips, denoted as $\textbf{A}_t^{prev}$ and $\textbf{V}_t^{prev}$. $\textbf{A}_t^{prev}$ and $\textbf{V}_t^{prev}$ are concatenated with the feature input $\textbf{A}_t$ and $\textbf{V}_t$. Position embeddings are added with the concatenated feature to identify the time order of the input tokens:
\begin{equation}
\begin{aligned}
    \tilde{\textbf{A}}_t = [\textbf{A}_t^{prev} \ \textbf{A}_t] + PE_a \\
    \tilde{\textbf{V}}_t = [\textbf{V}_t^{prev} \ \textbf{V}_t] + PE_v
\end{aligned}
\end{equation}
The audio and video features are passed through two separate MLP-Mixer blocks, Audio-Mixer and Video-Mixer. The two MLP-Mixer blocks here enable each modality to fuse information along the time dimension. 
The resulted audio and video features, $\hat{\textbf{A}}_t$ and $\hat{\textbf{V}}_t$, could capture the long-term feature from previous clips.
Then, we use the same strategy as visual Transformer~\cite{dosovitskiy2020image} by using a \texttt{[CLS]} token. 
\texttt{[CLS]} token is also a learnable vector concatenated with $\hat{\textbf{A}}_t$ and $\hat{\textbf{V}}_t$. The concatenated feature is then input to another MLP-Mixer block, Multimodal Mixer to mix information across different modalities. The output of \texttt{[CLS]} token, $\texttt{[CLS]}_t$ is used to represent the whole video clip with multimodal information. 
Finally, $\texttt{[CLS]}_t$ is concatenated with the input memory token $\texttt{[MEM]}_{t-1}$ and pass through another MLP-Mixer block, Memory-Mixer. 
In the Memory-Mixer, the current clip representation will interact with the memory token and update the memory content. The output at the position of $\texttt{[MEM]}_{t-1}$ is $\texttt{[MEM]}_{t}$, which is the input memory token for the next clip.
The memory output of the last clip $\texttt{[MEM]}_T$ will be used for scoring. 
Besides that, we also collect all the $\texttt{[CLS]}$ tokens output and add with position embeddings:
\begin{equation}
\begin{aligned}
    \tilde{\textbf{C}} & = [\texttt{CLS}_1 \ \texttt{CLS}_2 \  ... \ \texttt{CLS}_T] + PE_c 
\end{aligned}
\end{equation}
$\tilde{\textbf{C}}$ is then input to another Mixer block, CLS Mixer.
This block helps the model further learn the local information from each clip. The outputs of CLS Mixer are averaged and concatenated with $\texttt{[MEM]}_T$. The final score is generated by these two features with a linear layer.

\noindent\textbf{Skating Mixer \emph{vs} LSTM.}
The recurrent mechanism is similar to RNN~\cite{rnn} and LSTM~\cite{hochreiter1997long}, but the design of MLP-Mixer is simpler and is capable of dealing with multimodal data. In vanilla RNN architecture, when the input sequence becomes longer, gradient vanishing and gradient exploding happen.
In LSTM, this problem is solved by adding gate strategy. These gates control the information flow from the input to make the model focus on important parts of input and thus reduces the effective input length. In our architecture, it is not necessary to have delicately designed gates. Gradient vanishing and exploding issues could be mitigated since there is skip-connection within MLP-Mixer blocks and no extra projection is implemented for the memory token. Additionally, the proposed structure is capable of dealing with multimodal information, which is not considered in LSTM.
Mixing the memory token will enable the current clip to see the previous information and thus generate a comprehensive view for the whole video.

\noindent\textbf{Bi-direction Mixer.}
In our model, we use a similar strategy as in bidirectional recurrent neural network~\cite{rnn}, that is to add a backward direction in the model to grasp a comprehensive view of the whole video. For the backward direction, the last clip of the video will be processed first, then the memory flows to the first clip. 
The averaged value of $\texttt{[CLS]}$ outputs from forward and backward directions at each clip are input to CLS Mixer; the averaged $\texttt{[MEM]}$ output of the last clip in forward and backward directions will be used for scoring.
 

\vspace{-5pt}
\section{Experiment}

\begin{table*}[!t]
	\renewcommand\tabcolsep{5.4pt}
	\centering
	\small
	\scalebox{0.9}{
        \begin{tabular}{c|c|ccccccc|ccccccc}
        \toprule
        \rowcolor{blue!10}
         \multicolumn{1}{c|}{\textbf{Datasets}}&
         \multicolumn{1}{c|}{\textbf{Methods}} & &       &   \multicolumn{3}{c}{\textbf{\emph{Mean Square Error}($\downarrow$)}}   &  & &  & \multicolumn{5}{c}{\textbf{\emph{Spearman Correlation}($\uparrow$)}} &  \\ \cline{3-16}
         \rowcolor{blue!10}
          & 
          & \multicolumn{1}{c}{\textbf{\emph{TES}}}   &  \multicolumn{1}{c}{\textbf{\emph{PCS}}}     & \multicolumn{1}{c}{\textbf{\emph{SS}}}   &  \multicolumn{1}{c}{\textbf{\emph{TR}}}& \multicolumn{1}{c}{\textbf{\emph{PE}}}   &  \multicolumn{1}{c}{\textbf{\emph{CO}}}& \multicolumn{1}{c|}{\textbf{\emph{IN}}} & \multicolumn{1}{c}{\textbf{\emph{TES}}}   &  \multicolumn{1}{c}{\textbf{\emph{PCS}}}     & \multicolumn{1}{c}{\textbf{\emph{SS}}}   &  \multicolumn{1}{c}{\textbf{\emph{TR}}}& \multicolumn{1}{c}{\textbf{\emph{PE}}}   &  \multicolumn{1}{c}{\textbf{\emph{CO}}}& \multicolumn{1}{c}{\textbf{\emph{IN}}}  \\ \hline
       & C3D-LSTM & 39.25 & 21.97 & $\dagger$ & $\dagger$ & $\dagger$ & $\dagger$ & $\dagger$ & 0.29 & 0.51 & $\dagger$ & $\dagger$ & $\dagger$ & $\dagger$ & $\dagger$\\
      & MSCADC& 25.93 & 11.94 & $\dagger$ & $\dagger$ & $\dagger$ & $\dagger$ & $\dagger$ & 0.50 & 0.61 & $\dagger$ & $\dagger$ & $\dagger$ & $\dagger$ & $\dagger$\\ \cline{2-16}
      & M-LSTM & 25.70 & 12.48 & $\dagger$ & $\dagger$ & $\dagger$ & $\dagger$ & $\dagger$& 0.53 & 0.68 & $\dagger$ & $\dagger$ & $\dagger$ & $\dagger$ & $\dagger$\\
      & S-LSTM & 22.31 & 10.21 & $\dagger$ & $\dagger$ & $\dagger$ & $\dagger$ & $\dagger$& 0.57 & 0.74 & $\dagger$ & $\dagger$ & $\dagger$ & $\dagger$ & $\dagger$\\
      Fis-V& MS-LSTM & 22.64 & 9.84 & $\dagger$ & $\dagger$ & $\dagger$ & $\dagger$ & $\dagger$& 0.59 & 0.73 & $\dagger$ & $\dagger$ & $\dagger$ & $\dagger$ & $\dagger$\\ \cline{2-16}
      & M-BERT (Early)  & 28.04  &  13.31 & $\dagger$ & $\dagger$ & $\dagger$ & $\dagger$ & $\dagger$& 0.54 & 0.69 & $\dagger$ & $\dagger$ & $\dagger$ & $\dagger$ & $\dagger$  \\ 
      & M-BERT (Mid)  & 33.32 & 17.79 & $\dagger$ & $\dagger$ & $\dagger$ & $\dagger$ & $\dagger$ & 0.54 & 0.71 & $\dagger$ & $\dagger$ & $\dagger$ & $\dagger$ & $\dagger$        \\
      & M-BERT (Late)  & 27.73 & 12.38 & $\dagger$ & $\dagger$ & $\dagger$ & $\dagger$ & $\dagger$ & 0.53 & 0.72 & $\dagger$ & $\dagger$ & $\dagger$ & $\dagger$ & $\dagger$      \\  \cline{2-16}
      & Ours  &\bf 19.57 & \bf 7.96 & $\dagger$ & $\dagger$ & $\dagger$ & $\dagger$ & $\dagger$ & \bf 0.68 & \bf 0.82 & $\dagger$ & $\dagger$ & $\dagger$ & $\dagger$ & $\dagger$\\ \midrule
      & C3D-LSTM & 308.30 & 25.85 & 0.92 & 0.99 & 1.21 & 0.97 & 1.01 & 0.78 & 0.53 & 0.50 & 0.52 & 0.52 & 0.57 & 0.47\\
      & MSCADC & 148.02 & 15.47 & 0.51 & 0.57 & 0.78 & 0.55 & 0.60 & 0.77 & 0.70 & 0.69 & 0.69 & 0.71 & 0.68 & 0.71\\
        \cline{2-16}
      & M-LSTM  & 104.62 & 15.57 & 0.49 & 0.72 & 0.89 & 0.46 & 0.56 & 0.84 & 0.69 & 0.74 & 0.59 & 0.64 & 0.78 & 0.71  \\  
      & S-LSTM  & 83.79 & 10.90 & 0.40 & 0.43 & 0.70 & 0.40 & 0.44 & 0.87 & 0.79 & 0.79 & 0.80 & 0.78 & 0.80 &    0.80   \\  
      FS1000 (Ours) & MS-LSTM  & 94.55 & 11.03 & 0.45 & 0.49 & 0.76 & 0.43 & 0.47 & 0.86 & 0.80 & 0.77 & 0.78 & 0.76 & 0.79 & 0.78       \\  \cline{2-16}
      & M-BERT (Early)  & 139.09 & 14.49 & 0.44 & 0.44 & 0.71 & 0.47 & 0.50 & 0.78 & 0.77 & 0.80 & 0.80 & 0.76 & 0.79 & 0.79        \\ 
      & M-BERT (Mid)  & 170.57 & 21.28 & 0.57 & 0.54 & 0.69 & 0.56 & 0.56 & 0.77 & 0.75 & 0.79 & 0.79 & 0.80 & 0.79 & 0.80         \\ 
      & M-BERT (Late)  & 131.28 & 15.28 & 0.44 & 0.43 & 0.67 & 0.47 & 0.55 & 0.79 & 0.75 & 0.80 & 0.81 & 0.80 & 0.80  & 0.76         \\ \cline{2-16}
      & Ours   & \bf 81.24  & \bf 9.47 & \bf 0.35 & \bf 0.35 & \bf 0.62 & \bf 0.37 &   \bf 0.39 & \bf 0.88 & \bf 0.82 & \bf 0.80 & \bf 0.81 & \bf 0.80 & \bf 0.81 & \bf 0.81  \\ 
      \bottomrule
        \end{tabular}}
\vspace{-5pt}
\caption{Experiment Results on Fis-V~\cite{xu2019learning} and ours FS1000. \textbf{CNN-based}: \cite{parmar2017learning, parmar2019action}, \textbf{LSTM-based}:~\cite{xu2019learning,parmar2017learning}, \textbf{Transformer-based}:~\cite{lee2020parameter}, \textbf{MLP-based}: Ours. 
MSE and Spearman Correlation are used for evaluation. For MSE, the lower the better; for Spearman Correlation, the higher the better.
$\dagger$ denotes the dataset does not include the GT.}
\vspace{-15pt}
\label{tab: fs1000-reb}
\end{table*}




In this section, we show the experiment results on Fis-V dataset and FS1000 of different methods.
Fis-V dataset~\cite{xu2019learning} contains 400 ladies short program videos for training and 100 videos for validation. 
Our proposed FS1000 dataset contains a training set of 1000 videos and a validation set of 247 videos.
AST~\cite{gong2021ast} and TimeSformer~\cite{bertasius2021spacetime} are used as our feature extractor.
The Mean Square Error (MSE) and Spearman Correlation are used as our evaluation metrics. 
We compare our method with several related methods~\cite{xu2019learning,parmar2017learning,parmar2019action,lee2020parameter}.
M-Bert~\cite{lee2020parameter} uses both audio and visual features while others only consider visual features.
It should be noted that we find around 40 out of 500 videos in Fis-V~\cite{xu2019learning} dataset contains redundant information (such as the interview and replay), so we cut the videos and re-extract the features. 

\vspace{-2pt}
\subsection{Results on Fis-V}
From Table~\ref{tab: fs1000-reb}, it can be observed that our proposed Skating-Mixer outperforms CNN-based~\cite{parmar2017learning, parmar2019action}, LSTM-based~\cite{xu2019learning,parmar2017learning} and Transformer-based~\cite{lee2020parameter} models. Although the Transformer model performs better than MLP-Mixer on general tasks like image classification~\cite{tolstikhin2021mlp}, it does not have an obvious advantage in this specific task. C3D-LSTM~\cite{parmar2017learning} has the worst results since the model is too simple to learn such complex data. 3D CNN-based method, MSCADC~\cite{parmar2019action} is also struggled to understand such long videos. 
MS-LSTM~\cite{xu2019learning} and our proposed Skating Mixer performs better than Transformer model, indicating that: a strong memory mechanism plays an essential role in long videos learning; besides, the attention mechanism is less effective to capture extremely long-term dependencies across clips over several minutes. 
Additionally, by comparing our model and M-Bert~\cite{lee2020parameter}, it could be found that our model could better learn multimodal features from long videos.
\vspace{-2pt}
\subsection{Results on FS1000}
Scoring on the FS1000 dataset is much more challenging since Fis-V~\cite{xu2019learning} dataset only contains ladies short program videos while FS1000 dataset consists of different types of figure skating videos, which highly tests the robustness of model. 
As shown in Table~\ref{tab: fs1000-reb}, CNN-based~\cite{parmar2017learning,parmar2019action} and Transformer-based~\cite{lee2020parameter} models still cannot capture long-term dependencies across extremely long videos. 
LSTM-based~\cite{xu2019learning} models achieves better results but fails to further improve the results with only visual features.
Our proposed model could handle multimodal information in long figure skating videos and shows great flexibility to fit different types of competitions and obtain the best result among all the models.
Another interesting observation is that performance score \textbf{\emph{PE}} is harder to learn than other four sub-scores for all the models, indicating that future work could focus on improving the learning of performance score for a better overall result.


\vspace{-3pt}
\subsection{Ablation studies}
\begin{table*}[t]
\scriptsize
\centering
\scalebox{0.9}{
\begin{tabular}{lr|cccc|ccc|ccc}
\toprule
\rowcolor{blue!10}
\textbf{Factor} &
  \multicolumn{1}{l|}{} &
  \multicolumn{4}{c|}{\textbf{Component}} &
  \multicolumn{3}{c|}{\textbf{Modality}} &
  \multicolumn{3}{c}{\textbf{Scoring token}} \\
\rowcolor{blue!10}
\multicolumn{2}{l|}{}    & Mixer & Mixer+MEM & MRU & MRU+Bi-D & A & V & A+V & \texttt{[CLS]} & \texttt{[MEM]} & \texttt{[CLS]}+\texttt{[MEM]}\\ 
\hline
\multirow{2}{*}{\begin{tabular}[c]{@{}l@{}}Fis-V\\ \end{tabular}} &
  \textit{TES} &
  23.25 (0.55) &
  20.34 (0.66) &
  20.07 (0.66) &
  \bf 19.57 (0.68) &
  33.04 (0.49) &
  20.56 (0.67) &
  \bf 19.57 (0.68) &
  21.53 (0.66) &
  20.10 (0.67) &
  \bf 19.57 (0.68) 
 \\
 & \textit{PCS}   & 10.87 (0.73)  
 & 8.75 (0.79)
 & 8.23 (0.81)    
 & \bf 7.96 (0.82)     
 & 14.68 (0.67) 
 & 9.97 (0.76)
 & \bf 7.96 (0.82) 
 & 8.55 (0.81)
 & 8.10 (0.82)
 & \bf 7.96 (0.82)
 \\ \hline
\multirow{2}{*}{\begin{tabular}[c]{@{}l@{}}FS1000\\ (Ours)\end{tabular}} &
  \textit{TES} &
  93.65 (0.85) &
  90.80 (0.87) &
  85.71 (0.88) &
  \bf 81.24 (0.88) &
  94.94 (0.82) &
  82.59 (0.88) &
  \bf 81.24 (0.88) &
  83.16 (0.86)&
  88.78 (0.86)&
  \bf 81.24 (0.88)
  \\
 & \textit{PCS} & 13.81 (0.75) 
 & 10.96 (0.80) 
 & 10.44 (0.81)    
 & \bf 9.47 (0.82)   
 & 20.40 (0.55)
 & 10.92 (0.79) 
 & \bf 9.47 (0.82)  
 & 10.37 (0.81)
 & 10.71 (0.79)
 & \bf 9.47 (0.82) 
\\ \bottomrule
\end{tabular}
}
\vspace{-5pt}
\caption{Ablation studies on different components, modalities and scoring tokens. The value outside the brackets is MSE while inside is Spearman correlation.
}
\vspace{-10pt}
\label{fig:abl} \end{table*}

\begin{table}[h!]
\scriptsize
\centering
\scalebox{0.9}{
\begin{tabular}{l|c|cccc}
\toprule
\rowcolor{blue!10}
  \multicolumn{1}{c|}{\textbf{Dataset}} &
  \multicolumn{1}{c|}{\textbf{Backbone}} &
  \multicolumn{1}{c}{\textbf{\textit{TES}}} &
  \multicolumn{1}{c}{\textbf{\textit{PCS}}} &
  \multicolumn{1}{c}{\textbf{\# Params}} &
  \multicolumn{1}{c}{\textbf{MACs}}\\ \hline
\multirow{2}{*}{\begin{tabular}[c]{@{}l@{}}Fis-V\\ \end{tabular}} &
  Transformer & 25.60 (0.51)
  & 13.18 (0.68)
  & 20.48M
  & 35.55G
    \\
 & MLP-Mixer & \bf 19.57 (0.68) & \bf 7.96 (0.82) & \bf 14.32M & \bf 24.95G    \\ \hline
\multirow{2}{*}{\begin{tabular}[c]{@{}l@{}}FS1000\\ (Ours)\end{tabular}} &
  Transformer & 108.68 (0.81)
     & 13.40 (0.76)
     & 20.48M
     & 48.92G
  \\
 & MLP-Mixer & \bf 81.24 (0.88) & \bf 9.47 (0.82) & \bf 14.32M  & \bf 34.36G 
\\ \bottomrule
\end{tabular}
}
\caption{Comparison of Transformer and MLP-Mixer as backbone in our Skating Mixer. MACs means Multiply–Accumulate Operations.}
\vspace{-15pt}
\label{tab:backbone} \end{table}

We have conducted ablation experiments on both datasets. Here we analyze two major scores, \textbf{\emph{TES}} and \textbf{\emph{PCS}}.

\begin{figure}[t!]
\centering
\includegraphics[width=1.0\linewidth]{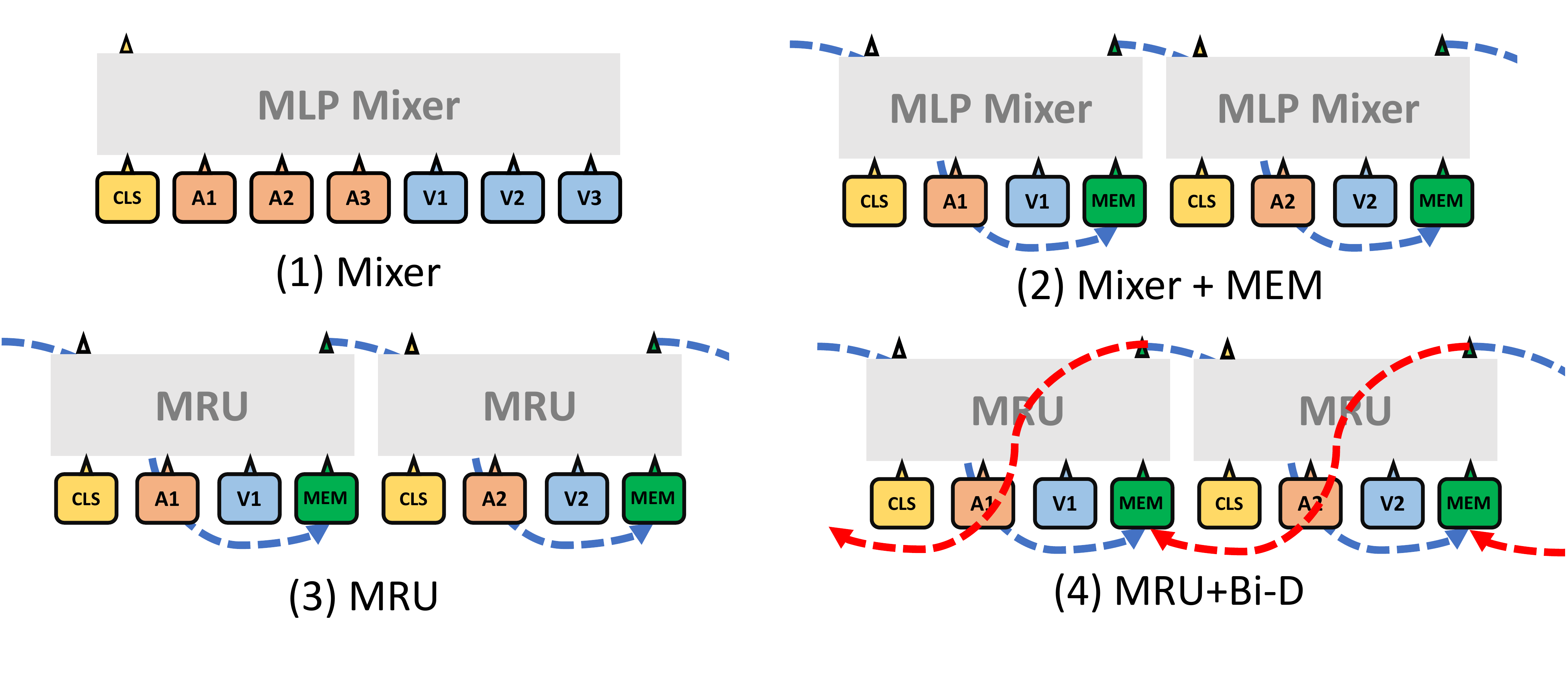}
\vspace{-20pt}
\caption{Four of our designed fusion structures. 
}
\vspace{-15pt}
\label{fig:abl_structure}\end{figure}

\noindent{\textbf{Component.}}
We first examine the effectiveness of our designed memory recurrent unit (MRU) and bi-direction propagation through four different fusion model structures, as shown in Figure~\ref{fig:abl_structure}. The first one (\textbf{Mixer}) is to simply input all the audio and video features into a large MLP-Mixer model for fusion. 
We then optimize such process by introducing the memory recurrent flow with MLP-Mixer (\textbf{Mixer+MEM}).
Then, the vanilla MLP-Mixer is replaced by our proposed MRU (\textbf{MRU}).
Finally, the bi-directional mechanism is applied with our proposed MRU (\textbf{MRU+Bi-D}).
The result is shown in Table~\ref{fig:abl}. 
The model fails to learn the long-term relationship between audio and video modalities when directly inputting all the features into a single model. Introducing memory mechanism helps the model capture long-term information and our proposed MRU could further better fuse multimodal features within extremely long videos. 
Moreover, adding bi-direction flow could effectively enhance the understanding of long videos and improve the performance of the model.

\noindent\textbf{Modality. }
Here we analyze the importance of using multimodal information in figure skating task. We have run experiments on our proposed model with only audio and only visual features. 
Results in Table~\ref{fig:abl} show that Skating Mixer with only audio features generates the worst result. 
This follows the commonsense that music plays an auxiliary part in figure skating and the visual technique moves always play the key role for scoring. 
Introducing both audio and video clues in figure skating scoring performs better than using only visual features, which further demonstrates that audio-visual learning is really important in this field.

\noindent\textbf{Scoring token. }
In this section, we are going to discuss the effectiveness of \texttt{[CLS]} and \texttt{[MEM]} token in our proposed architecture. In our model, both $\texttt{[CLS]}$ and $\texttt{[MEM]}$ token output are used for final scoring. We first generate scores with only the output from CLS Mixer, which means only \texttt{[CLS]} token outputs are used for scoring. Then we utilize only the \texttt{[MEM]} output from the last clip for scoring. The result show that using both $\texttt{[CLS]}$ and $\texttt{[MEM]}$ token generates the best result. 
This is because $\texttt{[MEM]}$ token passes across the whole video and contains global information while $\texttt{[CLS]}$ token contains local information of each clip. 
Combining these two types of features could improve the performance.

\noindent\textbf{Backbones.}
In this part, we show the difference between MLP-Mixer and Transformer.
Transformer has been widely adopted in the multimodal area, but as mentioned in~\cite{tolstikhin2021mlp}, the computation complexity of Transformer is $\mathcal{O}(n^2)$ while for MLP-Mixer it is $\mathcal{O}(n)$, where $n$ is the number of input tokens.
Also, \cite{Liu2021EfficientTO} mentions that Transformer usually requires large amount of data for training.
We conduct experiments by replacing the MLP-Mixer with the same number of Transformer blocks. 
The results in Table~\ref{tab:backbone} demonstrate that in our task with a relatively small dataset (compared to Imagenet~\cite{imagenet_cvpr09}), MLP-Mixer yields better performance than Transformer with less computation resource.

\subsection{Visualization}
\begin{figure}[h]
\begin{center}
\includegraphics[width=1.0\linewidth]{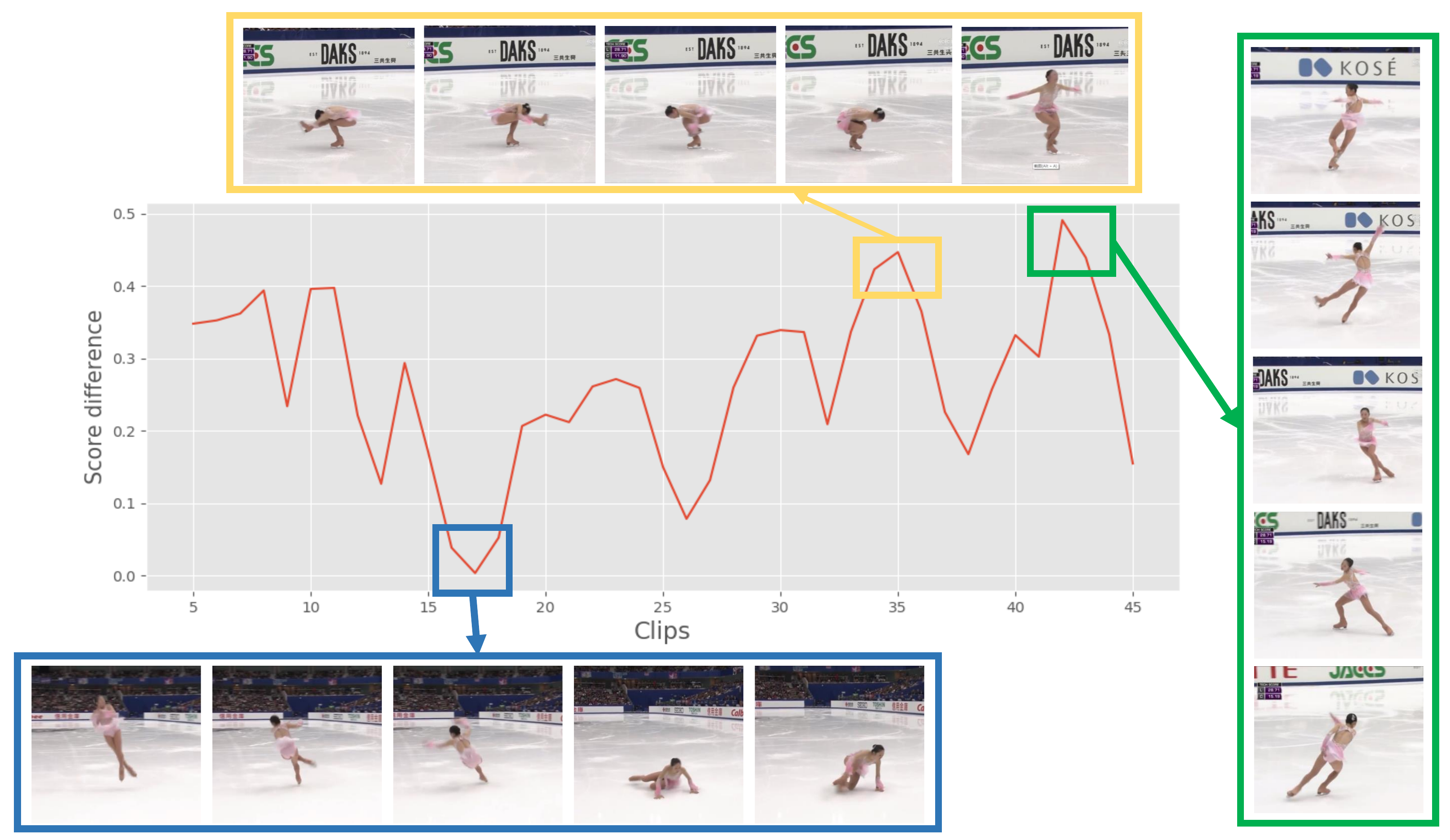}
\end{center}
\vspace*{-10pt}
    \caption{Score difference when sequentially adding clips. The score is low when the action is failed (blue box); the score is relatively high when finishing a complex move (yellow box) or skate fluently with the music (green box).}
\label{fig:visualization}
\vspace{-10pt}
\end{figure}
\noindent{In this section, we will demonstrate how our model scores in a certain video. We produce scores starting with the first clip and add one following clip at each time. When adding one clip, the score difference is computed to show the score of this clip. An example is shown in Figure~\ref{fig:visualization}. It can be seen that when an action is failed, the score for this clip is much lower (blue box). When the athlete successfully performs a complex technical move, the score is much higher (yellow box). Also, we found that when the athletes skate with the beat of the music, they also get higher scores (green box). This result clearly demonstrates that our proposed method could learn some basic patterns in figure skating.}

\vspace{-4pt}
\subsection{Performances on Beijing 2022 Olympic Games}
\begin{figure}[t]
\centering
\subfigure[Predicted results on Ladies Short Program]{
\includegraphics[width=\linewidth]{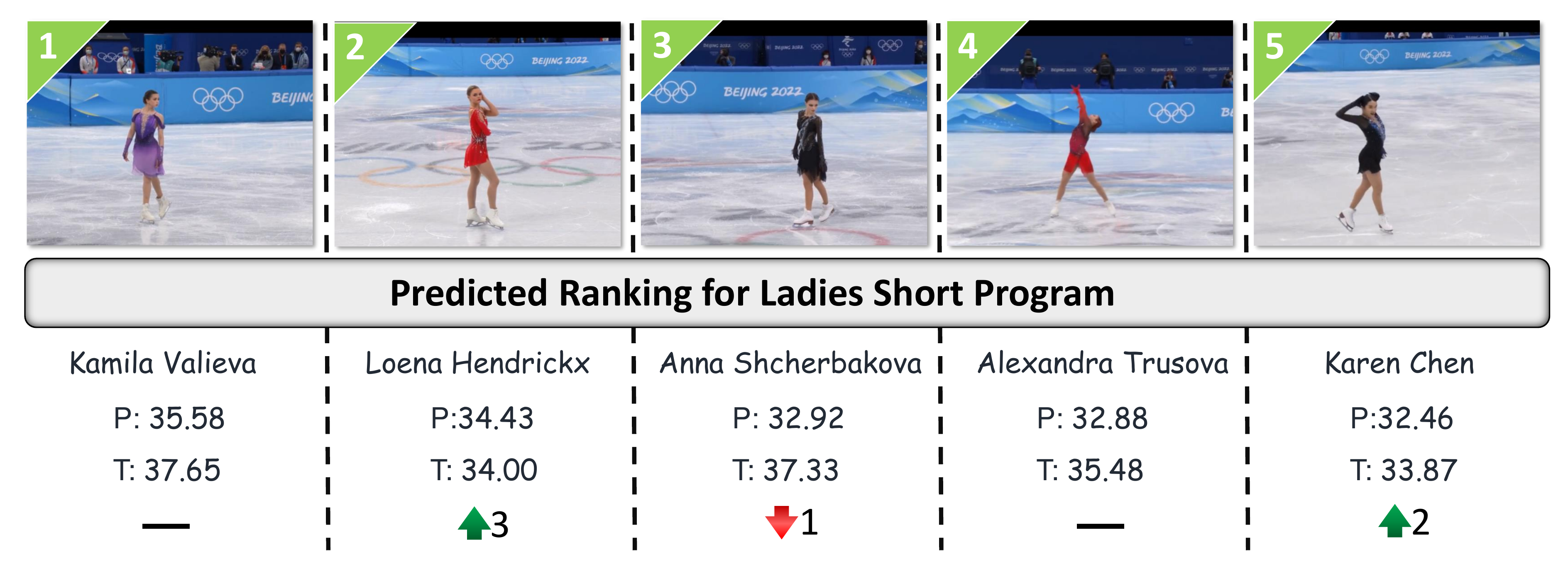}
}
\quad
\subfigure[Predicted results on Pairs Free Program]{
\includegraphics[width=\linewidth]{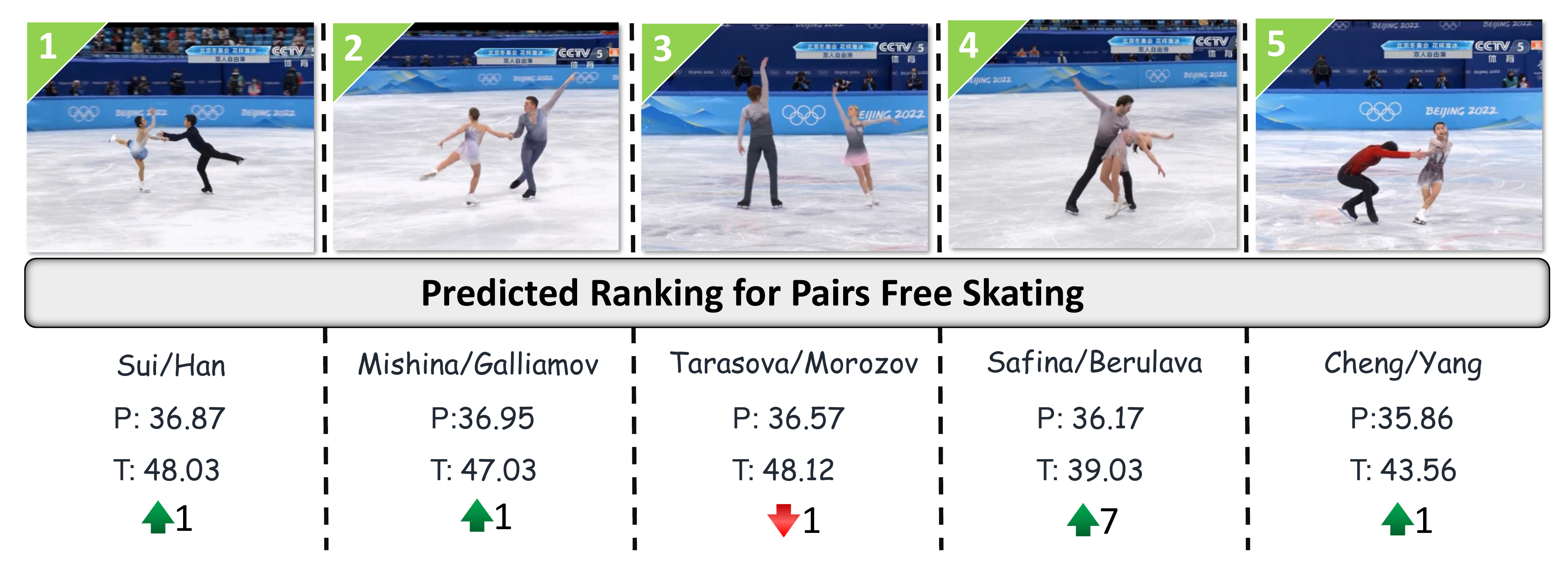}
}
\vspace{-5pt}
\caption{Predicted TOP-5 Ranking in Beijing 2022 Winter Olympic Games. $P$ stands for predicted score and $T$ stands for true score. The last row is the ranking difference compared to the real ranking.}
\label{fig:rank}
\vspace{-15pt}
\end{figure}

To verify the robustness and effectiveness of the figure skating model, 
we apply our trained model on competitions that are not included in the training data to see its actual effect. 
Therefore, we take the figure skating videos from the Beijing 2022 Winter Olympic Games and make analysis on the rankings of athletes. In practice, predicting correct ranking is more meaningful than predicting the exact scores because the evaluation scales are inconsistent on different competitions. Here we use our model to predict the \textbf{\emph{PCS}} for samples. Figure~\ref{fig:rank} demonstrates the results of Ladies Short and Pair Free program. It shows that although the score may not be accurate, the top-5 ranking does not change too much compared to real ranking. This is because top-tier athletes share similar technique moves and maintain high-quality performances. 
Such results could satisfy the practical need for auxiliary judgment.
This demonstrates that our model could actually learn some of the scoring standards and identify better performance from a group of athletes. 

\section{Related Work}

\subsection{Audio-Visual Learning}
There exists a rich exploration in audio-visual multimodal learning, especially in the deep learning era~\cite{zhu2021deep}. 
And datasets~\cite{chen2020vggsound,lee2021acav100m,gemmeke2017audio,abu2016youtube} involving representation learning in this field have been also developed.
%
%
%
M-BERT~\cite{lee2020parameter} focuses on reducing the complexity when introducing Transformer into audio-visual representation learning.
%
AVAS~\cite{morgado2020learning} attempts to learn the spatial alignment between audio and vision by introducing a new self-supervised proxy task.
Besides, another work~\cite{nagrani2021attention} also proposes to design a middle bottleneck to integrate audio and video modalities.
Although many existing methods explore audio-visual learning by various techniques, our proposed Skating-Mixer is the first attempt to apply MLP architecture to tackle this problem.

\subsection{MLP-based Architecture}
Attention-based network Transformer~\cite{vaswani2017attention,ji2022masked} achieve unparalleled success in NLP field, while convolutional networks~\cite{He_2016_CVPR} are still the main solution for most vision tasks~\cite{zhuge2022salient} until the proposal of ViT~\cite{dosovitskiy2020image}. 
Recently, MLP-Mixer~\cite{tolstikhin2021mlp} argue that MLP can be an alternative solution for vision representation learning.
Subsequently, many MLP-based architectures emerged in computer vision area.
S$^2$-MLP~\cite{yu2021s} only contains channel-mixing MLP and utilizes a spatial-shift operation for communication among patches. 
In addition, there are several specific-function models being proposed, such as CycleMLP~\cite{chen2021cyclemlp} used in dense prediction, MixerGAN~\cite{cazenavette2021mixergan} and CrossMLP~\cite{ren2021cascaded} in image translation, and Mixer-TTS~\cite{tatanov2021mixer} in the text-to-speech task. Distinguished from the aforementioned works, our proposed model is the first MLP model solving audio-visual problems.

\subsection{Figure Skating}
In the computer vision field, the earliest work about figure skating can be traced back to MIT-Skating~\cite{pirsiavash2014assessing}, which gathered Olympic videos and assessed the actions. 
Similar research~\cite{xu2019learning} also focuses on scoring figure skating videos and collects a Fis-V dataset with 500 videos. 
However, Fis-V dataset only considers ladies single program, making it hard for generalization in this field. 
\cite{liu2020fsd} introduces action recognition in figure skating and meanwhile designs an FSD-10 dataset. 
Another fine-grained, motion-centered MCFS dataset~\cite{DBLP:conf/aaai/LiuZLZXDZ21} is  proposed for the temporal action segmentation task. 
Additionally, several dedicated models have been proposed.
Specifically, \cite{nakano2020estimating} detects the highlight in figure skating programs with people's reaction.
ACTION-Net~\cite{zeng2020hybrid} strengthens the importance of postures in long videos, learning both context and temporal discriminative information. 
EAGLE-Eye~\cite{nekoui2021eagle} creates a two-stream pipeline to learn the long-term representation of figure skating actions. 
Compared with the others, our model is the first attempt to solve figure skating scenarios using both audio and video cues.

\section{Conclusion and Future Work}
This paper introduces the MLP-based multimodal architecture for scoring figure skating. 
The proposed model solves learning multimodal information in long videos, which is essential in this task.
Besides, an elaborated-designed dataset has been collected.
We set the benchmarks that compare our model with CNN-Based, LSTM-based and Transformer-Based methods on the Fis-V and our proposed FS1000 datasets.
The experiments show the effectiveness of our method, indicating MLP-based architecture is capable of the multimodal task. 
Furthermore, we apply our model on the 2022 Winter Olympics to verify the model's effectiveness and applicability.
\jf{However, this work mainly focuses on the figure skating. It would be an interesting point to extend the dataset and the method to other sports. Also, the pose information can be great supplementary to do action-based tasks, which can be used to improve the method in the future.}

\section{Acknowledgment}
This work was supported by the National Key R\&D Program of China (Grant NO. 2022YFF1202903) and the National Natural Science Foundation of China (Grant NO. 61972188 and 62122035).

\bibliography{main}

\end{document}


\maketitle
\section{MLP-Mixer}

Here we briefly introduce the MLP-Mixer module used in our model. In~\cite{tolstikhin2021mlp}, an MLP-based architecture has been proposed in the computer vision area and shows that 
model organizing multi-layer perceptrons are powerful enough to achieve excellent results on image classification tasks.
Similar to Visual Transformer~\cite{dosovitskiy2020image}, input images for MLP-Mixer~\cite{tolstikhin2021mlp} are split into several non-overlapping patches and each patch is treated as a sequence token. 
MLP-Mixer contains several identical layers, and each layer contains two MLP blocks: channel-mixing MLP and token-mixing MLP. 
Channel-mixing MLP operates on the channel dimension of the input feature, allowing different channels to communicate with each other; token-mixing MLP operates on the token dimension, so information could flow across different tokens and communicate with each other. 
Each block contains two MLP layers, and one GELU~\cite{hendrycks2020gaussian} activation function (described as $\Phi$). Besides, skip connection is also applied in each block. 

To be more specific, suppose $\mathbf{X} \in \mathbb{R}^{S \times C}$ is a two-dimension input feature, where $S$ is the sequence length (number of tokens) and $C$ is the number of channels for each token. In each layer, the function could be represented as:
\begin{equation}
\begin{aligned}
\mathbf{U}_{*, i}&=\mathbf{X}_{*, i}+\mathbf{W}_{2} \Phi \left(\mathbf{W}_{1} \text{Norm}(\mathbf{X}_{*, i})\right),\\
\mathbf{Y}_{j, *}&=\mathbf{U}_{j, *}+\mathbf{W}_{4} \Phi \left(\mathbf{W}_{3} \text{Norm}(\mathbf{U}_{j, *})\right),
\end{aligned}
\label{MLP-Mixer}
\end{equation}
where $i$ ranges from $1 \ldots C$, and $j$ ranges from $1 \ldots S$. 
$\text{Norm}$ denotes LayerNorm~\cite{ba2016layer} and $\mathbf{W}$ represents the weights of linear layer in each block. Input feature first passes through token-mixing MLP and then follows channel-mixing MLP.
This structure allows each element in the input feature could interact with other features along two dimensions.



\section{Feature Extraction Settings}
In our experiment, all the video data have 25 frames per second. Each video is separated into a 5-second clip and adjacent clips have 3 seconds overlapping time duration. The overlapping setting tends to avoid inconsistency caused by splitting. 
For feature extraction, we use the Audio Spectrogram Transformer~\cite{gong2021ast} pre-trained on full AudioSet~\cite{audioset} to extract acoustic features. For the visual feature, TimeSformer~\cite{bertasius2021spacetime} pre-trained on Kinetics-600 dataset~\cite{kay2017kinetics} is implemented. 
In the original paper, the number of input frames of TimeSformer is 8, so we also adopt the same setting here. 
To round up that 125 cannot be divided by 8, we take 120 frames from 125 frames for each clip.
So each clip is separated into 15 non-overlapping 8-frame segments and each segment is input into the model. 
In other words, there will be 15 tokens used as the visual representation for each clip. We do not fine-tune both Transformers on Fis-V and our dataset because of tremendous computational cost and memory usage. Figure skating videos contain up to 6000 frames and it will require $\sim$200G memory and more than 2 minutes for a single video to pass forward and backward the model. Thus they are not included in our training graph.

\section{Model Settings}
In our experiment, 
the number of layers in Video Mixer, Audio Mixer, Memory Mixer and CLS Mixer is 1 while the number of layers in Multimodal Mixer is 2. The input feature dimension is 512.
Adam optimizer with learning rate $1e-4$ and weight decay $5e-6$ is deployed. 
The whole framework is trained with 400 epochs for FisV dataset and 200 epochs for our proposed FS1000 dataset on a single NVIDIA V100. 

\section{Metrics}
In our experiment, we use Mean Square Error (MSE) and Spearman Correlation, which is commonly used in figure scoring task~\cite{xu2019learning, parmar2017learning}. Mean Square Error is a metric used to measure the value difference between the true score and the predicted score. Spearman correlation is a non-parametric metric to show the monotonicity of the relationship between two variables. In our case, we use this metric to calculate if the predicted results and the true results are similar to each other in ranking or not. Suppose the predicted score is $P$ and the true score is $T$. The number of samples is $N$. The equation of MSE could be written as:
\begin{equation*}
MSE=\frac{1}{N} ||P - T||^2_2
\end{equation*}
Smaller MSE indicates more accurate prediction.
To calculate the Spearman correlation, we first convert raw scores to rank $\mathrm{R}(P)$ and $\mathrm{R}(T)$.
The coefficient $r_{s}$ is computed as follow:
\begin{equation*}
r_{s}=\frac{\operatorname{cov}(\mathrm{R}(P), \mathrm{R}(T))}{\sigma_{\mathrm{R}(P)} \sigma_{\mathrm{R}(T)}}
\end{equation*}
A higher coefficient value means a better rank correlation between the true and predicted scores. 

\section{Results on Beijing Winter Olympics}
\begin{figure}[h!]
\centering
\subfigure[Predicted results on Ladies Short Program]{
\includegraphics[width=\linewidth]{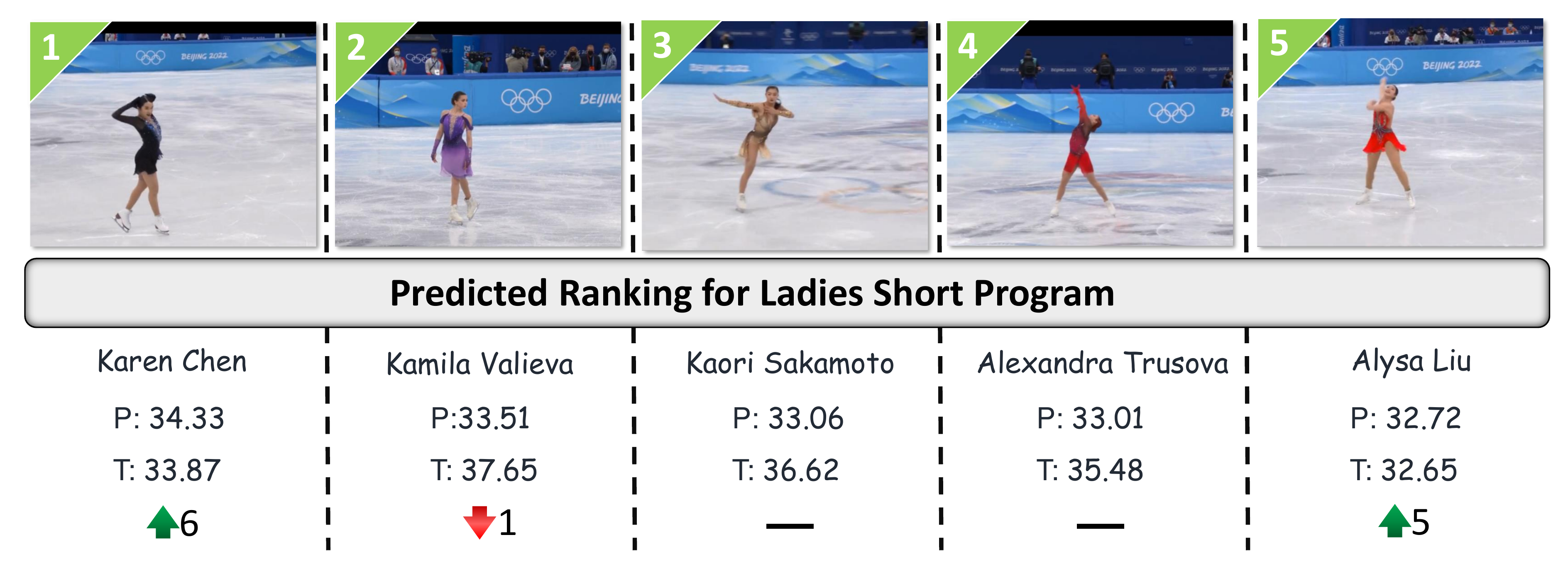}
}
\quad
\subfigure[Predicted results on Pairs Free Program]{
\includegraphics[width=\linewidth]{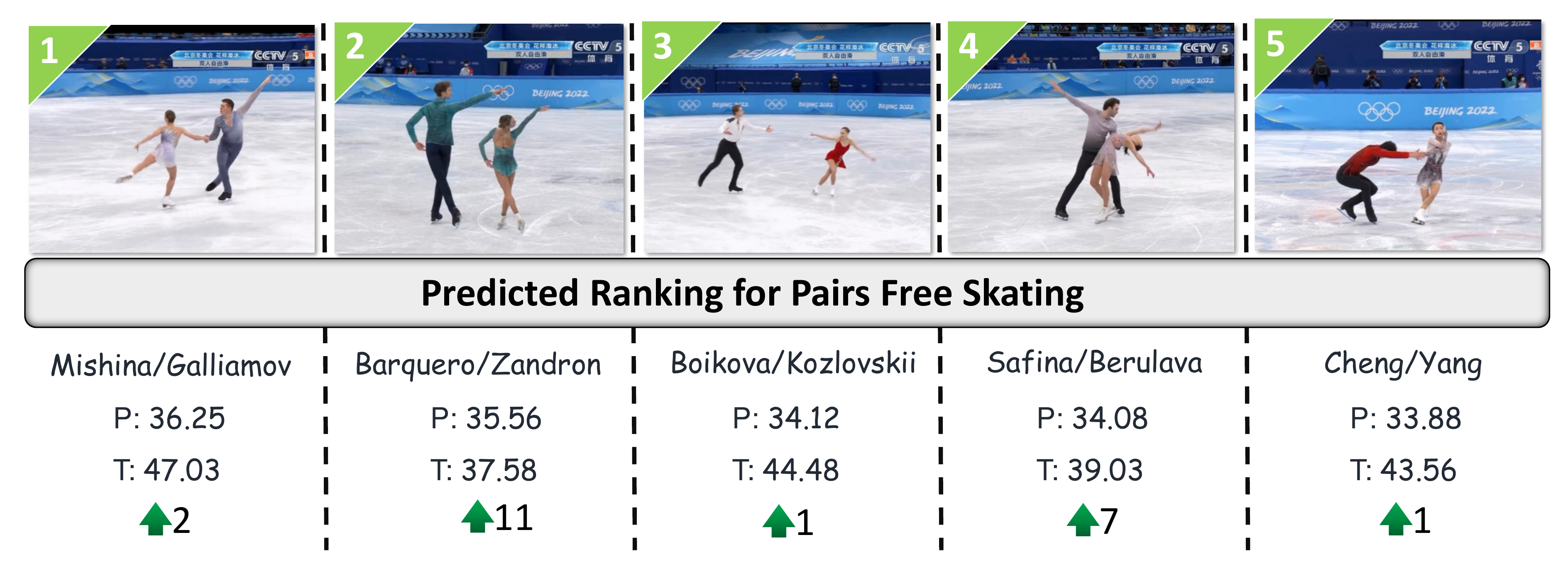}
}
\vspace{-5pt}
\caption{Predicted TOP-5 Ranking in Beijing 2022 Winter Olympic Games with S-LSTM~\cite{xu2019learning}.}
\label{fig:rank_slstm}
\end{figure}

\begin{figure}[h!]
\centering
\subfigure[Predicted results on Ladies Short Program]{
\includegraphics[width=\linewidth]{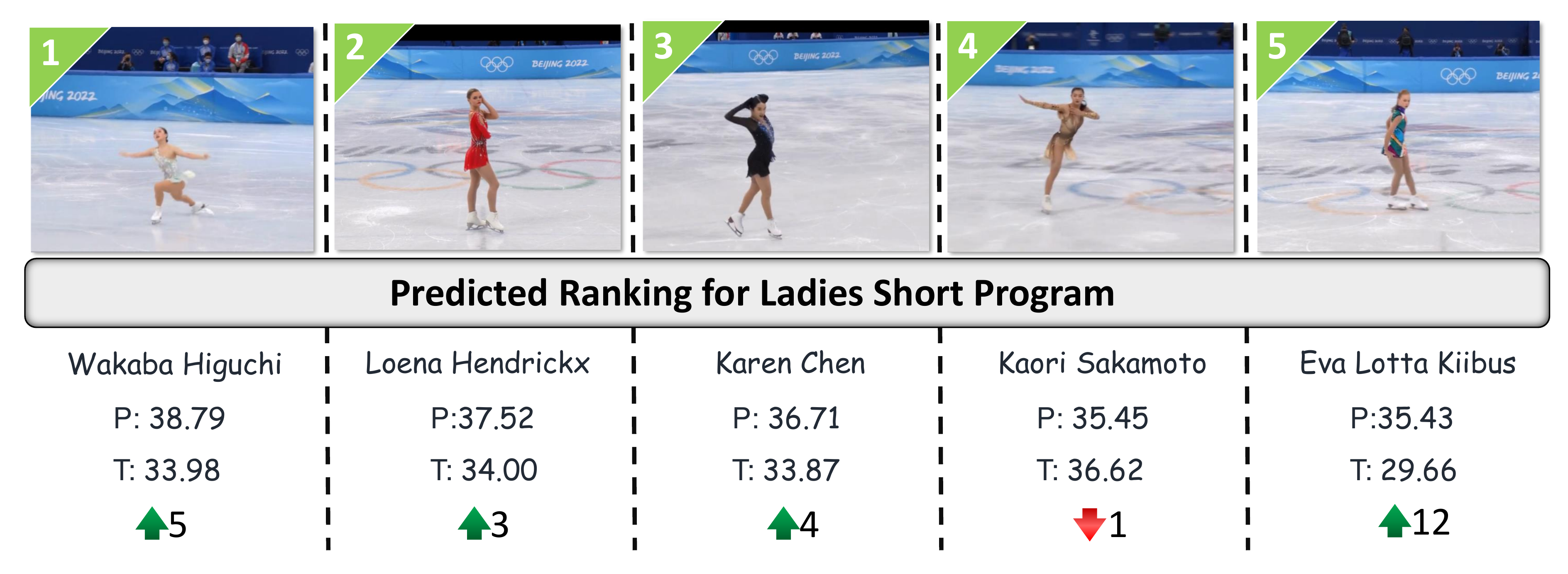}
}
\quad
\subfigure[Predicted results on Pairs Free Program]{
\includegraphics[width=\linewidth]{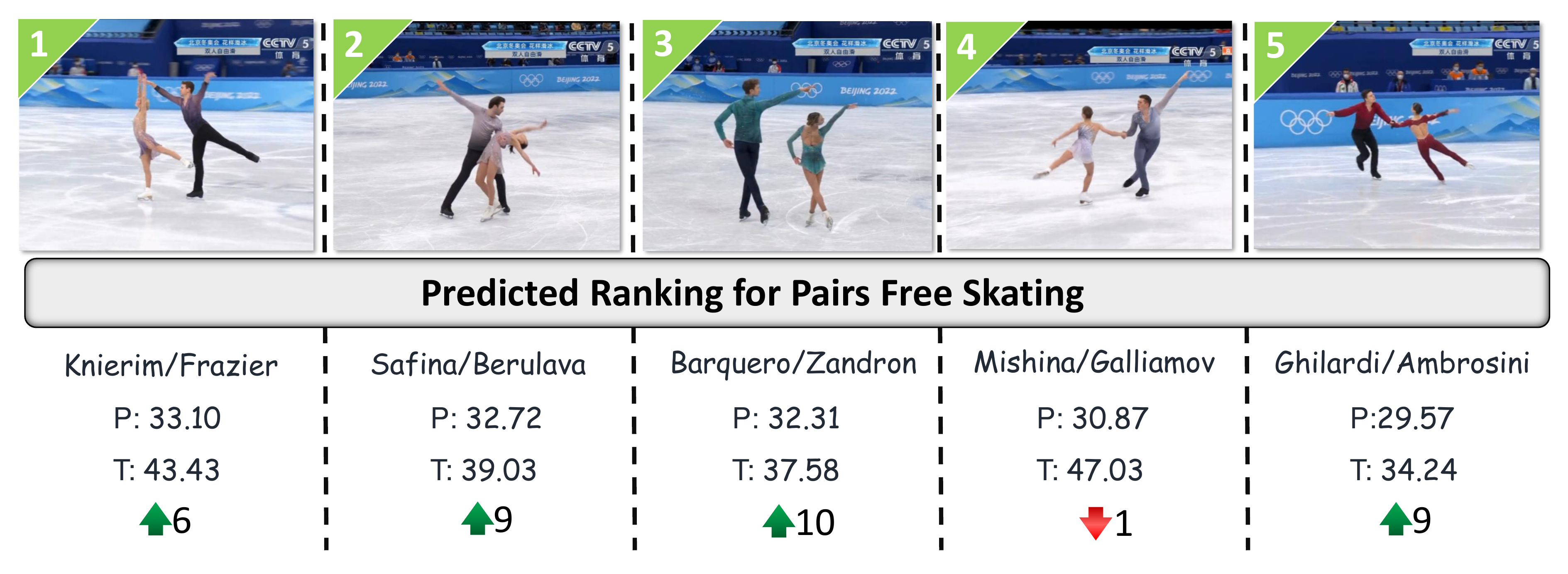}
}
\vspace{-5pt}
\caption{Predicted TOP-5 Ranking in Beijing 2022 Winter Olympic Games with MS-LSTM~\cite{xu2019learning}.}
\label{fig:rank_mslstm}
\end{figure}

In this part, we show more results of the Beijing 2022 Winter Olympic Games. In the section before, we present the ranking result of ladies short program and pairs free program using our proposed Skating Mixer. To show the advantages of our proposed model, here we choose S-LSTM and MS-LSTM~\cite{xu2019learning}, which have closer performance to our model as shown in the previous section, and apply these two models to the videos in Winter Olympic Games. The result is demonstrated in Figure~\ref{fig:rank_slstm} and ~\ref{fig:rank_mslstm}, respectively. 
We can see that the predicted ranking results of both models are much worse than those when using our proposed structure. 
We can also find that S-LSTM generates better results than MS-LSTM which follows the performance shown in the experiment section.
This again shows the models can capture some scoring rules and learn to assess figure skating programs. 
It also demonstrates that our proposed structure is much robust and could extend to other competitions that are not within the training dataset.

\subsection{Visualizations}
\begin{figure}[h!]
\centering
\subfigure[Visualization on Men Free Program]{
\includegraphics[width=\linewidth]{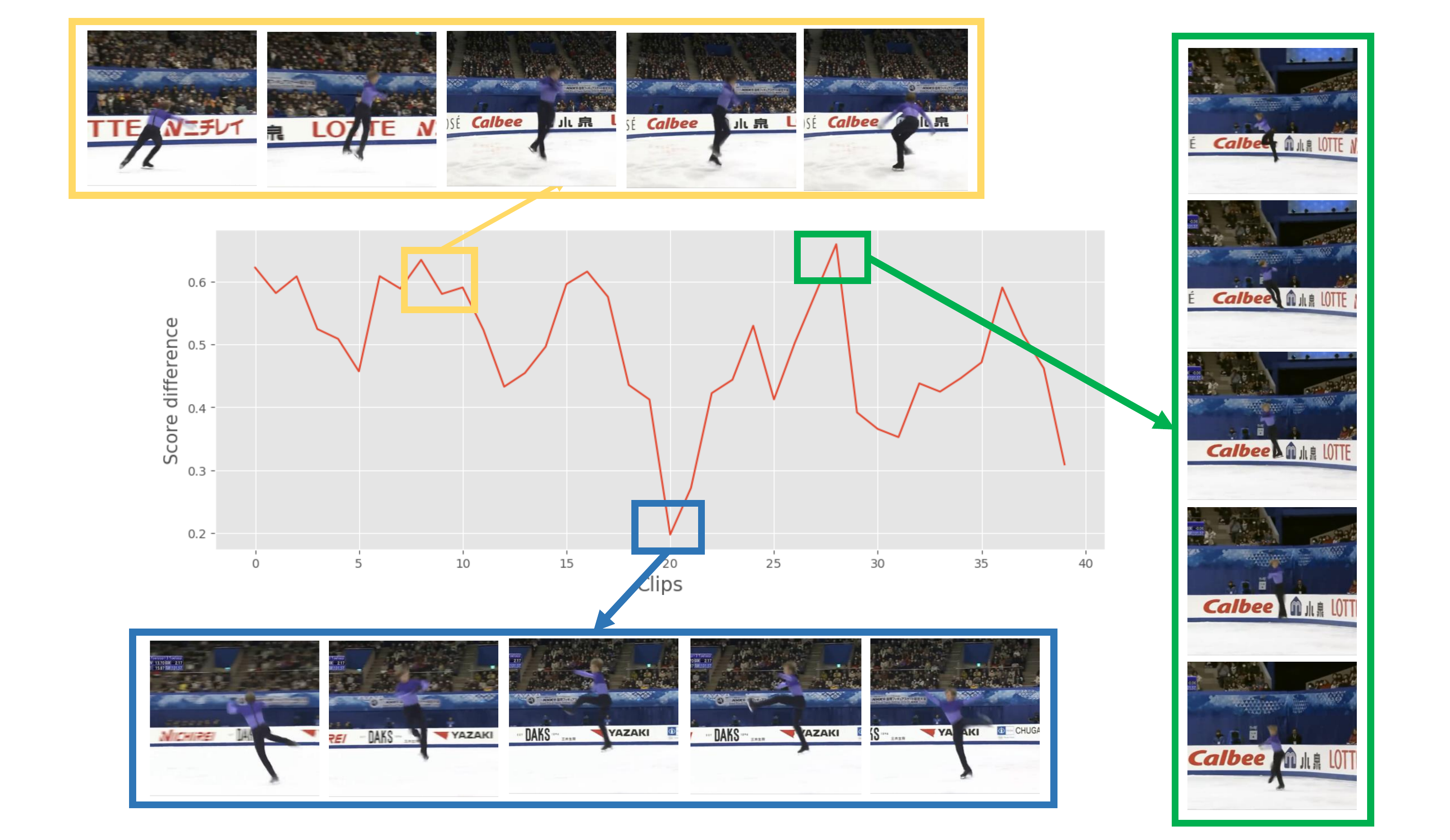}
}
\quad
\subfigure[Visualization on Pairs Free Program]{
\includegraphics[width=\linewidth]{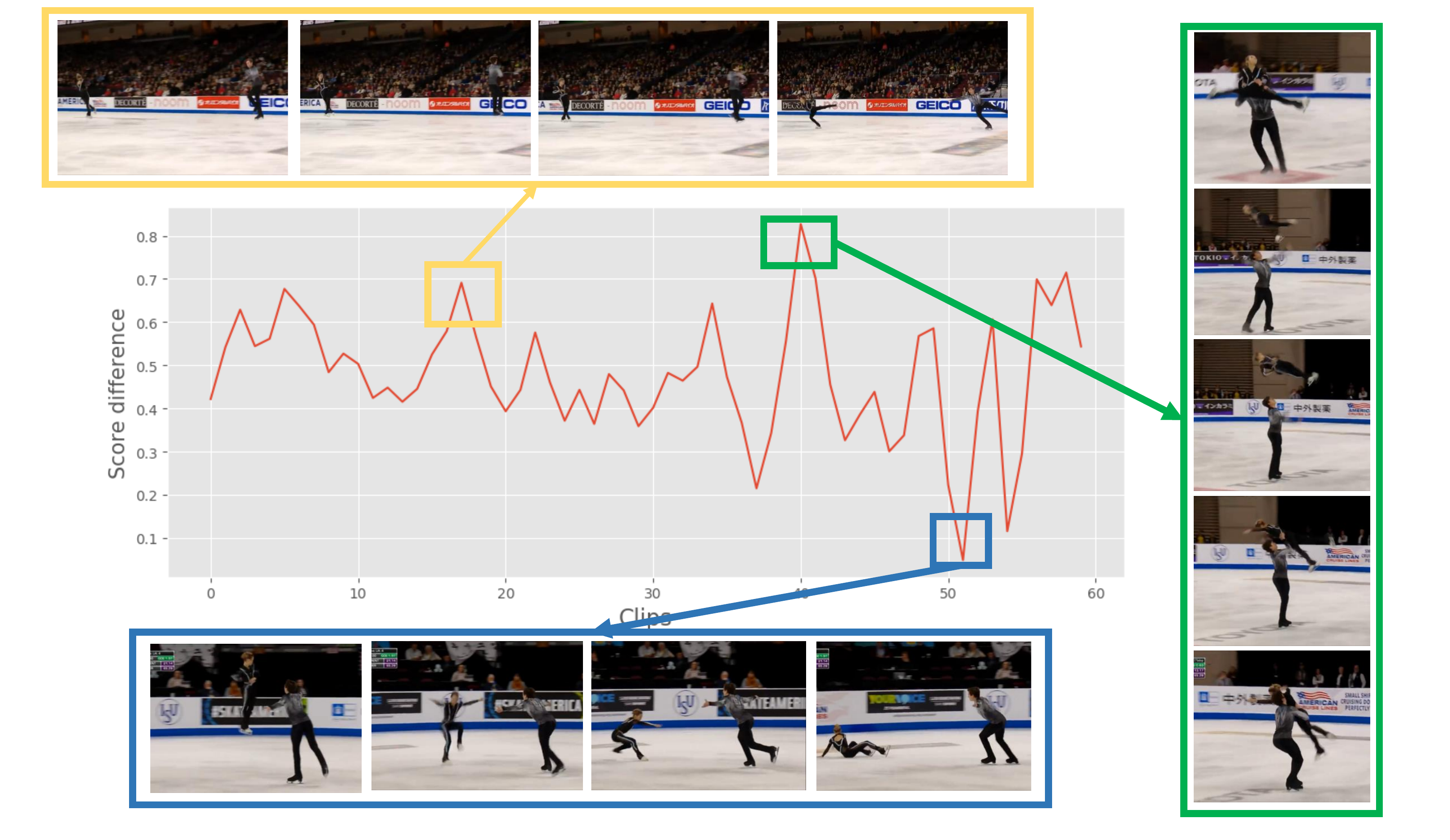}
}
\vspace{-5pt}
\caption{Score difference when sequentially adding clips. The blue box indicates the failed action; the yellow and green box show the successful and fluent actions. }
\label{fig:visualize2}
\vspace{-15pt}
\end{figure}

In this part, we show more examples of visualizations as mentioned in the previous section. Figure~\ref{fig:visualize2} are two examples from men free program and pairs free program. We can see that when the action is failed, the score for the corresponding clip is relatively low; when the action is successfully done or the move is elegant and fluent, the score is relatively high. In addition, in the pairs free program, the action synchronization is also important for scoring. This implies that our model can learn some basic standards in figure skating scoring. 

\section{Core Code}
In this section, we show the core code implemented in our structure.
Listing~\ref{list:mlp_block} demonstrates the code of our proposed memory recurrent unit (MRU). The input memory token is first input to two bottleneck structures to get the information of each modality in previous clips. Then it is concatenated with the feature in the current clip. The features passes through audio mixer, video mixer and multimodal mixer. Finally the output of \texttt{[CLS]} token is mixed with \texttt{[MEM]} in the memory mixer. The \texttt{[MEM]} output is used for the next clip and the \texttt{[CLS]} token outputs are also collected for scoring.
\begin{lstlisting}[language=Python, caption=Code of MRU, label={list:mlp_block}, float=*]
def model_forward(self, audio_feature, video_feature, hidden_state=None, back=False):

        if hidden_state is None:
            hidden_state = self.initial_memory

        prev_audio = self.audio_extraction(hidden_state)
        prev_video = self.video_extraction(hidden_state)

        if back:
            concat_audio_input = torch.cat([audio_feature, prev_audio], dim=1)
            concat_video_input = torch.cat([video_feature, prev_video], dim=1)
        else:
            concat_audio_input = torch.cat([prev_audio, audio_feature], dim=1)
            concat_video_input = torch.cat([prev_video, video_feature], dim=1)
            
        audio_idx = torch.arange(audio_len + 1)
        video_idx = torch.arange(video_len + 1)

        audio_pos_encoding = self.audio_pos_embedding(audio_idx)
        video_pos_encoding = self.video_pos_embedding(video_idx)

        audio_input = concat_audio_input + audio_pos_encoding
        video_input = concat_video_input + video_pos_encoding

        audio_output = self.audio_mixer(audio_input)
        video_output = self.video_mixer(video_input)
        
        cls_token = self.class_token

        multimodal_input = torch.cat([cls_token, audio_output, video_output], dim=1)
        multimodal_output = self.multimodal_mixer(multimodal_input)

        out_cls = multimodal_output[:, 0:1]

        memory_input = torch.cat([hidden_state, out_cls], dim=1)
        memory_output = self.memory_mixer(memory_input)

        out_hs = memory_output[:, 0:1]

        return out_hs, out_cls
\end{lstlisting}


\bibliography{aaai23}